\documentclass[11pt]{article}

\IfFileExists{acl.sty}{%
  \usepackage[preprint]{acl}%
}{%
  \usepackage{natbib}%
}
\usepackage{times}
\usepackage{latexsym}
\usepackage[T1]{fontenc}
\usepackage[utf8]{inputenc}
\usepackage{microtype}
\usepackage{amsmath}
\usepackage{amssymb}
\usepackage{booktabs}
\usepackage{multirow}
\usepackage{graphicx}
\usepackage{placeins}
\usepackage{capt-of}
\usepackage{needspace}
\IfFileExists{acl.sty}{%
  \newenvironment{widetable}{\begin{table*}[!t]}{\end{table*}}%
}{%
}

\title{Beyond Scores and Rankings: Diagnostic Fidelity for Efficient Multilingual Evaluation}

\author{
  Chenxin Liu\textsuperscript{1} \quad
  Yilun Liu\textsuperscript{1} \quad
  Zhenzhen Qin\textsuperscript{1} \quad
  Haofeng Li\textsuperscript{1} \quad
  Miao Zhang\textsuperscript{1} \quad
  Chen Liu\textsuperscript{1} \\
  \textbf{Shimin Tao}\textsuperscript{1} \quad
  \textbf{Li Zhang}\textsuperscript{1} \quad
  \textbf{Xiaojun Meng}\textsuperscript{1} \quad
  \textbf{Liqun Deng}\textsuperscript{1} \quad
  \textbf{Daimeng Wei}\textsuperscript{1} \\
  \textsuperscript{1}Huawei Technologies Co., Ltd., China \\
  \texttt{liuchenxin1@huawei.com}
}

\begin{document}
\maketitle

\begin{abstract}
Developing multilingual large language models requires repeated evaluation
across languages and capabilities, making comprehensive benchmarking costly.
Benchmark compression reduces this cost by estimating full-benchmark
performance from a small subset of examples, but preserving overall scores
and model rankings does not guarantee a reliable diagnosis of each model's
weak capabilities. We propose \emph{diagnostic fidelity} as a complementary
criterion, measured by profile alignment and weakness recovery, and find
that the methods that best preserve rankings do not necessarily best
preserve diagnosis. We then introduce \emph{Diagnostic Anchor} (DiagAnchor),
which allocates queries across capability groups, selects representative
items by source-model response patterns, and calibrates predictions for
unqueried items from each target model's answers on these anchors.
Averaged over eight multilingual benchmarks, DiagAnchor attains the highest
profile alignment at query budgets of 1\%, 5\%, and 10\%, and improves
weakness Recall@3 over the strongest baseline by 3.9 and 9.0 percentage
points at 5\% and 10\%. At a 5\% query budget, DiagAnchor already exceeds
the best baseline at 10\% in both full-profile alignment and weakness
Recall@3. DiagAnchor also outperforms the baselines on both diagnostic
metrics at 5\% and 10\% when earlier models serve as sources for newer
targets.
\end{abstract}

\section{Introduction}
\label{sec:introduction}

Developing large language models (LLMs) requires repeated evaluation as
training data, optimization procedures, and checkpoints evolve
\citep{biderman2023pythia}. In multilingual settings, this evaluation must
cover many tasks across many languages, which makes it particularly
costly. Benchmark compression reduces this cost by evaluating a model
on a small subset of items and estimating its full-benchmark performance.
Existing methods select representative items and estimate scores with
item response theory \citep{polo2024tinybenchmarks}, adapt subsets to
individual target models \citep{yuan2025beyond}, or directly optimize
for ranking preservation \citep{saranathan2025sublime}. These methods
are primarily assessed by \emph{score fidelity}, agreement with
full-benchmark scores, and \emph{ranking fidelity}, agreement with
full-benchmark model rankings. We refer to them jointly as
\emph{aggregate fidelity}. Both support model comparison and
checkpoint selection.

Model development also uses evaluation to guide data collection,
mixture adjustment, and the detection of capability-specific
regressions. For example, after
observing weaknesses in mathematics, the OLMo~2 developers used a
mathematics development set to guide mid-training data selection
\citep{olmo20242}. Such decisions depend on a model's profile across
capabilities, a within-model quantity that neither criterion directly
measures: score fidelity concerns agreement in aggregate scores, while
ranking fidelity compares models rather than capabilities. A compressed
benchmark can therefore preserve both while distorting the profile, for
example by retaining few items from a small, weak group. In our
experiments at a 10\% query budget, uniform random sampling achieves a
ranking correlation of 0.94 but recovers only 41\% of each model's three
weakest capability groups on average. tinyBenchmarks attains the highest
ranking correlation among the compared methods (0.96), yet still misses
about one of these three groups per model. Aggregate metrics alone
therefore provide limited evidence that compression preserves diagnosis.

We introduce \emph{diagnostic fidelity}, the extent to which compressed
evaluation preserves a model's capability profile and the weaknesses
identified by full evaluation. Anchor Points uses instance-level
reconstruction to examine model strengths and weaknesses
\citep{vivek2024anchor}. Building on this diagnostic use of compression,
we define two group-level criteria. \emph{Profile alignment}, the
Spearman correlation between full and estimated group scores, measures
preservation of relative capability strengths. \emph{Weakness recovery},
Recall@$k$ over the $k$ lowest-scoring groups, measures whether
compression identifies the same capabilities for potential improvement.
Both criteria apply to any evaluator-specified grouping. We instantiate
them using M-Tag, a shared multilingual taxonomy that serves as a proxy
for capability grouping.

Diagnostic groups are known before evaluation, but global representative
selection does not explicitly ensure their coverage.
We propose \emph{Diagnostic Anchor} (DiagAnchor), which uses these groups
to allocate queries and the target's responses to calibrate predictions.
First, each group receives at least one query when the
budget permits, and the remaining budget is distributed by power
allocation \citep{bankier1988power} to balance group coverage and group
size. Within each group, representative items, or \emph{anchors}, are
selected by clustering source-model responses. Second, calibration
weights source models by their agreement with the target on these
anchors. Each unqueried item is matched to an anchor in the resulting
response space. The difference between the target's observed anchor
score and the source-weighted prediction is then added to the item's
own source-weighted prediction.

Our contributions are as follows:
\begin{itemize}
    \item We formalize diagnostic fidelity as a within-model criterion
    for benchmark compression and show that aggregate and diagnostic
    fidelity favor different methods: at 5\% and 10\% budgets, the
    method with the highest ranking fidelity attains neither the
    highest profile alignment nor the highest weakness recall.

    \item We propose DiagAnchor, combining capability-stratified anchor
    selection with target-adaptive local calibration. Averaged over
    eight multilingual benchmarks, it attains the highest values on
    both diagnostic metrics at 5\% and 10\% budgets, improving
    Recall@3 over the strongest baseline by 3.9 and 9.0 percentage
    points, respectively. These advantages persist under a temporal
    holdout with source models that predate the targets according to
    the recorded release dates.

    \item Through ablation and sensitivity analyses, we show how the
    benefits of selection and calibration vary with the budget:
    stratified selection yields larger weakness-recovery gains over
    global selection as the budget grows, calibration helps weakness
    recovery most when queries are scarce, and the allocation exponent
    shifts the balance between diagnostic and aggregate fidelity.
\end{itemize}

\section{Related Work}
\label{sec:related_work}
\subsection{Efficient LLM Evaluation and Benchmark Compression}

Reducing evaluation cost requires understanding which information can
be retained when only a fraction of a benchmark is evaluated.
Active testing addresses a related annotation-cost problem by adaptively
selecting test items to label for performance estimation
\citep{kossen2021active}.
\citet{perlitz2024efficient} investigate this trade-off by examining
how benchmark design decisions affect the reliability of model
comparisons, showing that substantial reductions in evaluation cost
can be achieved while retaining reliable rankings. Subsequent work
has developed more targeted subset selection strategies. For example,
SubLIME predicts the ranking fidelity of candidate subsets from
anchor-model responses and dataset characteristics, and selects
subsets expected to preserve full-benchmark model rankings
\citep{saranathan2025sublime}.
In computer vision, Sort \& Search reuses historical model evaluations
to select test samples for approximate accuracy estimation as benchmarks
grow \citep{prabhu2024lifelong}.

A related line of research exploits response patterns across previously
evaluated models to select informative examples and estimate performance
on unobserved items. Anchor Points clusters
items by correlations in source-model
predictions and uses representative items to predict responses on the
rest \citep{vivek2024anchor}. Earlier studies use item response theory
to characterize item informativeness and the ability of test sets to
distinguish models \citep{rodriguez2021evaluation,vania2021comparing}.
tinyBenchmarks combines representative
selection with item response theory \citep{polo2024tinybenchmarks},
and \citet{pacchiardi2024instances} train an assessor that predicts
instance-level performance from responses on a small reference set and
instance features. TailoredBench further adapts estimation to each
target model by identifying the source models whose responses are most
consistent with it \citep{yuan2025beyond}. These methods are primarily
evaluated by aggregate fidelity and generally select items without
regard to predefined capability groups. DiagAnchor builds on
response-based selection and estimation, but uses the diagnostic
grouping to allocate queries and evaluates whether compression
preserves group-level conclusions.
% VERIFY: "generally select items without regard to predefined capability
% groups" -- check that tinyBenchmarks does not stratify by subject in
% the configurations you compare against.
 
\subsection{Diagnostic and Multilingual Evaluation}
 
Diagnostic evaluation characterizes model behavior beyond aggregate
performance. CheckList organizes behavioral tests around linguistic
capabilities \citep{ribeiro2020checklist}, and HELM evaluates models
across a taxonomy of scenarios and metrics \citep{liang2023helm}. In
multilingual evaluation, XTREME measures cross-lingual generalization
across languages and tasks \citep{hu2020xtreme}, and GaoYao organizes
evaluation into cultural layers and cognitive sub-layers
\citep{liu2026gaoyao}. Because these frameworks define groupings for
different purposes, their groups are not directly comparable across
benchmarks; we therefore use M-Tag as a shared capability label space
for defining diagnostic groups (Section~\ref{sec:diagnostic-space}).
Within benchmark compression, Anchor Points uses reconstructed
instance-level predictions to visualize regions of model strength and
weakness \citep{vivek2024anchor}. We instead assess whether compression
preserves conclusions over predefined groups, through profile alignment
and weakness recovery.

\section{Methodology}
\label{sec:methodology}
We first define capability groups and formulate two metrics for
diagnostic fidelity. We then introduce DiagAnchor, which selects
representative items within these groups and uses the target's
responses to calibrate item-level predictions.

\subsection{Diagnostic Fidelity: Formulation and Measurement}
\subsubsection{A Unified Diagnostic Space}
\label{sec:diagnostic-space}
Capability diagnosis requires a grouping of benchmark items by the
abilities they assess. Native benchmark labels differ in availability
and organizing principle, so we construct \emph{M-Tag}, a shared
taxonomy with 35 level-1 capability labels and 227 nested level-2
labels. A multilingual tagger assigns these labels, achieving level-1
accuracy of 87.5\% on human-annotated in-domain examples and 80.0\%
on out-of-domain examples.
Appendix~\ref{app:mtag} describes the taxonomy, the construction of the
training data, and the evaluation of annotation quality.
 
We use M-Tag as an operational proxy for capability rather than a
definitive account of it. Each level-1 category defines a capability
group, pooling items across languages. Level-2 categories are not used
because many categories contain too few items within a single benchmark. Profiles
and fidelity metrics are computed separately within each benchmark.
The formulation below also applies to other evaluator-specified
capability groupings.
 
\subsubsection{Formulating and Measuring Diagnostic Fidelity}
\label{sec:diagnostic_fidelity}

For a benchmark with item set $\mathcal{I}$ and $N=|\mathcal{I}|$,
let $\mathcal{G}$ denote its nonempty capability groups, defined here
by level-1 M-Tag labels. For a given target model, $y_i\in[0,1]$
denotes its score on item $i$, and $\mathcal{I}_g$ contains the items
assigned to group $g\in\mathcal{G}$. The full-benchmark group score is
\begin{equation}
    \mu_g = \frac{1}{|\mathcal{I}_g|}
    \sum_{i \in \mathcal{I}_g} y_i.
\end{equation}
Let $\widehat{\mu}_g$ denote the corresponding estimate obtained
from a compressed evaluation. Using the same group ordering,
the full and estimated capability profiles are
\begin{equation}
    \mathbf{p} = (\mu_g)_{g \in \mathcal{G}},
    \qquad
    \widehat{\mathbf{p}}
    = (\widehat{\mu}_g)_{g \in \mathcal{G}}.
\end{equation}
We assess diagnostic fidelity through two complementary
dimensions: preservation of the relative performance profile
and recovery of the model's weakest capability groups.

\paragraph{Profile alignment.}
Profile alignment measures whether compression preserves the
relative ordering of group-level performance within a model.
We quantify it using Spearman's rank correlation:
\begin{equation}
    \mathrm{PA}
    = \operatorname{Corr}\!\left(
        \operatorname{rank}(\mathbf{p}),
        \operatorname{rank}(\widehat{\mathbf{p}})
      \right),
\end{equation}
where $\operatorname{Corr}$ denotes Pearson correlation
between the rank vectors, with average ranks assigned to tied
scores. A higher value indicates better preservation of the
model's relative strengths and weaknesses across groups.
The metric is undefined when either rank vector is constant.

\paragraph{Weakness recovery.}
Weakness recovery measures whether compression identifies the
groups in which the model performs worst under full evaluation.
For a chosen $h$ with $1 \leq h < |\mathcal{G}|$, let
$\mathcal{W}_h$ contain the $h$ groups with the lowest
full-benchmark scores $\mu_g$, and let
$\widehat{\mathcal{W}}_h$ contain the $h$ groups with the lowest
estimated scores $\widehat{\mu}_g$.
We define
\begin{equation}
    \mathrm{Recall}@h
    = \frac{
        |\mathcal{W}_h \cap \widehat{\mathcal{W}}_h|
      }{h}.
\end{equation}
A higher value indicates that more of the weaknesses identified
by full evaluation are recovered by the compressed evaluation.
This metric focuses on the weakest groups, complementing
profile alignment, which considers the ordering of all groups.

% Both metrics are computed for each target model over the capability
% groups within a benchmark. Full-benchmark target scores provide the
% evaluation reference; only scores on queried items are available to
% the compressed estimator.

\subsection{Diagnostic Anchor}
\label{sec:diaganchor}
\begin{figure*}[t]
    \centering
    \includegraphics[width=\textwidth]{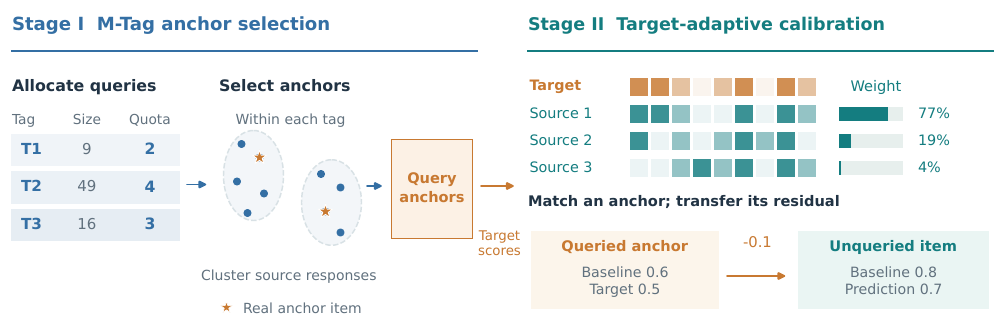}
    \caption{Overview of DiagAnchor's selection and calibration stages.
    Stars mark anchors; darker cells indicate higher scores.
    Numerical examples are illustrative.}
    \label{fig:diagnostic-anchor}
\end{figure*}

Let $x_{im}\in[0,1]$ be the score of source model $m$ on item $i$,
for $m=1,\ldots,M$. Given complete source responses and a query
budget $K$, DiagAnchor estimates the target's overall score and
capability profile using only $K$ target-item observations
(Figure~\ref{fig:diagnostic-anchor}).

\subsubsection{Stage I: Capability Anchor Selection}
\label{sec:anchor_selection}
Drawing on the principle of stratified sampling, we treat each
capability group used for diagnostic evaluation as a separate
stratum in selection. Indexing the $H=|\mathcal{G}|$ nonempty M-Tag groups
by $h$, we define
$\mathcal{I}_h=\{i\in\mathcal{I}:t_i=h\}$ and
$N_h=|\mathcal{I}_h|$, where $t_i$ is the item's level-1 M-Tag.
Items from different languages are pooled within each stratum.

\paragraph{Budget allocation.}
We assign each group an allocation weight
\begin{equation}
    w_h = N_h^\beta,
\end{equation}
where $\beta$ interpolates between equal group weights ($\beta=0$)
and size-proportional weights ($\beta=1$). When $K\geq H$, we
reserve one query per group and distribute the remaining budget
according to $w_h$. Otherwise, we sample $K$ distinct groups
without replacement according to these weights, assigning one
query to each. This prioritizes group coverage before allocating
additional observations. Integer allocation with group-size limits
is specified in Appendix~\ref{app:budget-allocation}.

\paragraph{Behavioral clustering and anchor selection.}
Let $k_h$ denote the quota for group $h$. Within each group receiving
queries, we cluster standardized source-response vectors using KMeans
and select the actual item nearest each cluster center. When response
patterns repeat, we adapt the cluster count and number of representatives
to fill the quota (Appendix~\ref{app:anchor-implementation}). The selected
items form a fixed anchor set $S_T$ with $|S_T|=K$.

Selection uses only item labels and source scores, allowing the anchors
to be reused across target models for a fixed source pool, budget, and
seed. All $K$ queries can therefore be issued without waiting for
intermediate target responses.

\subsubsection{Stage II: Target-Adaptive Local Calibration}
\label{sec:local_calibration}

We query the target on $S_T$ and use the observed scores
$\{y_j:j\in S_T\}$ to construct and locally calibrate a source-based
prediction for each item.

\paragraph{Source weighting.}
For each source model $m$, let $d_m$ denote its mean squared
score discrepancy from the target over the queried anchors.
We convert these distances into source weights $\alpha_m$
using a temperature $\tau>0$, and construct a source-weighted
prediction baseline $a_i$ for each item:
\begin{equation}
    \begin{aligned}
        d_m &= \frac{1}{K}\sum_{j\in S_T}(x_{jm}-y_j)^2,\\
        \alpha_m &= \frac{\exp(-d_m/\tau)}
            {\sum_{q=1}^{M}\exp(-d_q/\tau)},\\
        a_i &= \sum_{m=1}^{M}\alpha_m x_{im}.
    \end{aligned}
\end{equation}
Sources with smaller discrepancies receive higher weights.
The same source weights are used across all items.

\paragraph{Representative anchor matching.}
Let $z_{im}$ denote source scores standardized across the item pool
within each source dimension. We match each unqueried item to an
anchor using a target-weighted Manhattan distance:
\begin{equation}
    \begin{aligned}
        D_{\alpha}(i,j) &= \sum_{m=1}^{M}\alpha_m |z_{im}-z_{jm}|,\\
        r(i) &= \arg\min_{j\in S_T}D_{\alpha}(i,j).
    \end{aligned}
\end{equation}
Matching considers all selected anchors, including those from other
languages or M-Tag groups. Standardization and matching conventions
are given in Appendix~\ref{app:calibration-implementation}.

\paragraph{Local residual transfer.}
The residual $e_j=y_j-a_j$ captures the target's deviation from
the source baseline at anchor $j$. We assume that items matched to the same anchor have similar
prediction errors. We therefore apply the anchor's observed
residual as a shared correction to these items:
\begin{equation}
    \widetilde{y}_i
    = a_i + e_{r(i)}
    = a_i + y_{r(i)} - a_{r(i)}.
\end{equation}
Equivalently, the prediction starts from the target's
observed anchor score and adds the source-estimated score
difference between the current item and its anchor.

We retain the observed scores for queried items and project
unqueried predictions onto the valid score range:
\begin{equation}
    \widehat{y}_i =
    \begin{cases}
        y_i, & i\in S_T,\\
        \min\{1,\max\{0,\widetilde{y}_i\}\}, & i\notin S_T.
    \end{cases}
    \label{eq:bounded-prediction}
\end{equation}

\paragraph{Score and profile reconstruction.}
We aggregate the item-level predictions over the full benchmark
and within each capability group $g\in\mathcal{G}$:
\begin{equation}
    \widehat{\mu}
    = \frac{1}{N}\sum_{i\in\mathcal{I}}\widehat{y}_i,
    \qquad
    \widehat{\mu}_g
    = \frac{1}{|\mathcal{I}_g|}
      \sum_{i\in\mathcal{I}_g}\widehat{y}_i.
\end{equation}
These group estimates form the capability profile. Groups with no
queried items can still be estimated through reconstruction, without
requiring additional target queries.

\section{Experiments}
\label{sec:experiments}
\subsection{Experimental Setup}
\label{sec:experimental_setup}

\paragraph{Benchmarks and diagnostic groups.}
We evaluate on eight multilingual benchmarks spanning
reading comprehension, knowledge and reasoning,
mathematical problem solving, instruction following,
multi-turn dialogue, and cultural understanding.
Appendix Table~\ref{tab:benchmarks} summarizes the number
of evaluation items, languages, and M-Tag categories in
each benchmark. Diagnostic groups are defined by level-1 M-Tag
labels from the shared capability taxonomy described in
Section~\ref{sec:diagnostic-space}, pooling items across languages.

\paragraph{Source--target splits.}
Each benchmark's pool contains 40--42 model configurations. We use three splits, each
with disjoint source and target sets at approximately a 7:3 ratio, and
five selection seeds, shared across methods. Source models provide complete item-level
scores; target scores are accessible only through budgeted queries,
with full responses retained solely as evaluation references.
Appendix~\ref{app:model_splits} describes the split protocol, and
Table~\ref{tab:benchmarks} gives benchmark-specific model counts.

\paragraph{Baselines and query budgets.}
We compare Diagnostic Anchor with two main types of baselines: 
\begin{itemize}
    \item sampling-based baselines: uniform random sampling and M-Tag stratified sampling;
    \item model-based baselines: Anchor Points (APP, score adaptation)
\citep{vivek2024anchor}, tinyBenchmarks~\citep{polo2024tinybenchmarks},
and TailoredBench \citep{yuan2025beyond}.
\end{itemize}
We evaluate query budgets of $1\%$, $5\%$, and $10\%$
of each benchmark's evaluation pool. All methods are
compared under the same total number of target-item
queries, including any preliminary probes used for
target adaptation. Source-model scores are treated
as pre-existing reference data and do not count
toward the target query budget.

\paragraph{Evaluation metrics and aggregation.}
We report M-Tag profile Spearman correlation and weakness Recall@3,
alongside overall score MAE and model-ranking Spearman correlation.
Profile is reported as \emph{Full (Common)}: Full uses all M-Tag
groups within each benchmark and averages each method's defined
complete profiles. Since sampling-based baselines can leave some
groups without estimates, Common compares methods on their shared
M-Tag groups and evaluations.
Other metrics use matched evaluations across methods.
We average over target models, splits, and seeds, giving contributing
benchmarks equal weight and excluding undefined values.

\paragraph{Implementation details.}
For Diagnostic Anchor, we use a budget allocation
exponent of $\beta=0.5$ and a source-weighting temperature
of $\tau=0.1$. We examine sensitivity to $\beta$ and $\tau$ in
Section~\ref{sec:sensitivity}.
Clustering settings, numerical conventions, and
tie-breaking rules are provided in
Appendix~\ref{app:implementation_details}.

\subsection{Main Results: Aggregate and Diagnostic Fidelity}
\label{sec:main-results}

\begin{table*}[t]
\centering
\small
\setlength{\tabcolsep}{4pt}
\renewcommand{\arraystretch}{1.06}
\begin{tabular}{@{}llrrrr@{}}
\toprule
& & \multicolumn{2}{c}{Diagnostic fidelity} & \multicolumn{2}{c}{Aggregate fidelity} \\
\cmidrule(lr){3-4}\cmidrule(lr){5-6}
Budget & Method & \shortstack{Profile $\rho\uparrow$\\Full (Common)}
& Recall@3 $\uparrow$ & Ranking $\uparrow$ & MAE $\downarrow$ \\
\midrule
1\% & Random & \textsc{na} (0.2806) & 0.2333 & 0.8390 & 0.0458 \\
 & M-Tag Random & 0.3856 (0.2947) & 0.3016 & 0.8056 & 0.0720 \\
 & Anchor Points (APP) & 0.6761 (0.6863) & \textbf{0.6069} & \textbf{0.8851} & 0.0312 \\
 & TailoredBench & 0.6595 (0.6953) & 0.5905 & 0.8573 & 0.0312 \\
 & tinyBenchmarks (C, gpIRT) & 0.6384 (0.6358) & 0.5807 & 0.8747 & 0.0341 \\
 & tinyBenchmarks (I, gpIRT) & 0.6573 (0.6174) & 0.5812 & 0.8453 & \textbf{0.0311} \\
 & \textbf{DiagAnchor (ours)} & \textbf{0.7069} (\textbf{0.7203}) & 0.5969 & 0.8356 & \textbf{0.0311} \\
\midrule
5\% & Random & \textsc{na} (0.4029) & 0.3647 & 0.9222 & 0.0194 \\
 & M-Tag Random & 0.6308 (0.5085) & 0.4571 & 0.8845 & 0.0260 \\
 & Anchor Points (APP) & 0.6825 (0.7266) & 0.6207 & 0.9286 & 0.0216 \\
 & TailoredBench & 0.6729 (0.7237) & 0.6154 & \textbf{0.9306} & 0.0220 \\
 & tinyBenchmarks (C, gpIRT) & 0.6848 (0.7499) & 0.6211 & 0.9273 & 0.0203 \\
 & tinyBenchmarks (I, gpIRT) & 0.6852 (0.7440) & 0.6160 & 0.9067 & \textbf{0.0180} \\
 & \textbf{DiagAnchor (ours)} & \textbf{0.8414} (\textbf{0.7772}) & \textbf{0.6597} & 0.9102 & 0.0198 \\
\midrule
10\% & Random & \textsc{na} (0.4932) & 0.4140 & 0.9375 & 0.0125 \\
 & M-Tag Random & 0.7812 (0.6648) & 0.5693 & 0.9321 & 0.0158 \\
 & Anchor Points (APP) & 0.6925 (0.7426) & 0.6334 & 0.9349 & 0.0190 \\
 & TailoredBench & 0.7060 (0.7607) & 0.6314 & 0.9442 & 0.0173 \\
 & tinyBenchmarks (C, gpIRT) & 0.7084 (0.7745) & 0.6251 & \textbf{0.9588} & 0.0130 \\
 & tinyBenchmarks (I, gpIRT) & 0.6982 (0.7658) & 0.6233 & 0.9393 & \textbf{0.0116} \\
 & \textbf{DiagAnchor (ours)} & \textbf{0.8962} (\textbf{0.8269}) & \textbf{0.7231} & 0.9364 & 0.0172 \\
\bottomrule
\end{tabular}
\caption{Diagnostic and aggregate fidelity across eight benchmarks.
Bold marks the best mean, including ties. C/I: correctness/IRT embeddings;
APP: score-adapted Anchor Points; NA: undefined mean.}
\label{tab:main-results}
\end{table*}

Table~\ref{tab:main-results} reports score, ranking, and diagnostic fidelity;
tinyBenchmarks' pIRT variants are reported in
Appendix~\ref{app:pirt-results}.

\paragraph{DiagAnchor improves capability-level diagnostic fidelity at 5\% and 10\% budgets.}
DiagAnchor achieves the highest mean profile correlation under both
Full and Common evaluation, as well as the highest weakness Recall@3,
at these budgets. Full-profile correlations reach 0.841 and 0.896,
compared with the strongest baselines' 0.685 and 0.781, while
Recall@3 exceeds the strongest baseline by 3.9 and 9.0
percentage points, respectively.

\paragraph{Aggregate and diagnostic fidelity tend to favor different methods.}
Uniform random sampling already attains competitive aggregate fidelity
at 5\% and 10\%, despite substantially lower diagnostic fidelity.
At 10\%, its ranking correlation is 0.0213 below the best method,
yet it recovers only 41\% of the weakest capability groups.
The same divergence appears among model-based methods: at both budgets,
the method with the highest ranking correlation attains neither the
highest profile alignment nor the highest Recall@3.
DiagAnchor's diagnostic gains also involve an aggregate-fidelity
trade-off. At 5\% and 10\%, its ranking correlations are about 0.02
below the best method, and its MAEs are comparable to those of TailoredBench and
Anchor Points but exceed the lowest errors.

At 1\%, DiagAnchor leads in profile alignment, while Anchor Points
leads in weakness recovery and model ranking. Their ranking correlations
are 0.8356 and 0.8851, respectively, a gap of 0.0495.

% \paragraph{The smallest budget yields mixed diagnostic outcomes.}
% At 1\%, DiagAnchor has the highest average Full and Common profile
% correlations (0.707 and 0.720), exceeding the strongest baseline
% under each protocol by 0.0308 and 0.025, respectively.
% However, its Recall@3 of 0.6 trails Anchor Points' 0.61 by
% 1.00 percentage point. Anchor Points also achieves the highest ranking
% correlation (0.885), exceeding DiagAnchor's 0.836 by about 0.05.
% DiagAnchor's MAE of 0.0311 ties the lowest value at the displayed
% precision. Improved profile alignment therefore does not imply the
% best weakness recovery or model-ranking fidelity at this budget.

% \paragraph{Diagnostic improvements vary by benchmark and profile protocol.}
% Appendix Tables~\ref{tab:per-benchmark-mtag-1}
% and~\ref{tab:per-benchmark-mtag-2} show that DiagAnchor achieves
% the highest Full-profile correlation on all eight benchmarks at
% both 5\% and 10\%. Under Common Profile, it leads on five
% benchmarks at 5\% and seven at 10\%; the exceptions are INCLUDE,
% MGSM, and MMMLU at 5\%, and SAGE at 10\%.
% For Recall@3, it leads on four benchmarks at 5\% and all eight
% at 10\%. At 1\%, the advantage is less widespread: DiagAnchor
% leads on four benchmarks under Full Profile, three under Common
% Profile, and one for Recall@3. These differences indicate that
% the reliability of a compressed diagnosis depends on both the
% benchmark and the aspect of diagnostic fidelity being measured.

\paragraph{The advantage becomes more consistent across benchmarks as the budget grows.}
DiagAnchor achieves the highest Full-profile correlation on all eight
benchmarks at both 5\% and 10\%. Appendix~\ref{app:diagnostic-results}
reports the benchmark-specific results, including the more mixed outcomes
under Common Profile, Recall@3, and the 1\% budget.

At 5\% and 10\% budgets, DiagAnchor also leads both language-level
diagnostic metrics, despite using no language-based stratification
(Appendix~\ref{app:language-diagnostics}).

\subsection{Generalization to Newer Models}
\label{sec:generalization}
In iterative model development, responses from previously evaluated
models are reused to assess new checkpoints. To approximate this setting,
we hold out the ten most recent dated model configurations and use
earlier models from other checkpoint groups as sources. Each benchmark
uses a fixed pool of 25--27 sources, depending on response availability.
Appendix~\ref{app:temporal-protocol} describes the source eligibility
criteria and temporal split protocol.

We use the baselines from Section~\ref{sec:main-results}, with
tinyBenchmarks restricted to its two gpIRT variants.
Table~\ref{tab:logo-best-baseline} compares DiagAnchor with the best
baseline for each metric. At 5\% and 10\%, DiagAnchor again leads
on all diagnostic measures, exceeding the best baseline by 0.146
and 0.089 in Full-profile correlation and by 1.8 and 4.8 percentage
points in Recall@3. Although its Full-profile correlation and
Recall@3 are lower than under random splits at these budgets,
its advantage over the baselines is preserved.
As in the main results, DiagAnchor retains the highest profile
correlations at 1\% but trails in Recall@3, and its MAE exceeds
the lowest baseline error at 5\% and 10\%.
Ranking fidelity leads narrowly at 5\% but trails the strongest
baseline at the other budgets, with the largest gap at 1\%.

% We use the same baselines as in \ref{sec:main-results}. Table~\ref{tab:logo-best-baseline} reports Full and
% Common Profile, Recall@3, and MAE against the best baseline for each
% metric and budget. DiagAnchor achieves the highest mean Full and Common Profile
% correlations at all three budgets. At 5\% and 10\%, its Full-profile
% correlations reach 0.8216 and 0.8730, exceeding the strongest baseline
% by 0.1457 and 0.0888, respectively. The corresponding Common-profile
% gains are 0.0587 and 0.1094, showing that the advantage also holds on
% the shared diagnostic groups. Weakness Recall@3 reaches 0.6297 and
% 0.6991, improving on the strongest baseline by 1.79 and 4.81 percentage
% points. Thus, at these budgets, the advantage extends to both
% capability ordering and identification of the weakest groups.

% The results also show the limits of this advantage. At 1\%, DiagAnchor
% has the lowest MAE and highest profile correlations, but its Recall@3
% of 0.5732 is below the best baseline's 0.5937. At 5\% and 10\%, its
% MAEs of 0.0203 and 0.0194 exceed the best baseline's 0.0145 and 0.0103.
% The LOGO results therefore provide further evidence of diagnostic
% benefits at moderate query budgets, together with a trade-off in
% aggregate-score estimation and less consistent weakness recovery at
% the smallest budget.

\begin{table}[t]
\centering
\small
\setlength{\tabcolsep}{1.5pt}
\renewcommand{\arraystretch}{1.12}
\begin{tabular}{@{}llrrrr@{}}
\toprule
Budget & Method & Profile $\rho\uparrow$ & R@3$\uparrow$ & Rank$\uparrow$ & MAE$\downarrow$ \\
 & & Full (Common) & & & \\
\midrule
% 1% best baseline for MAE: Anchor Points / APP, 0.0344.
% 1% best baseline for Full Profile: Anchor Points / APP, 0.6525.
% 1% best baseline for Common Profile: Anchor Points / APP, 0.6715.
% 1% best baseline for Recall@3: tinyBenchmarks correctness + gpIRT, 0.5937.
% 1% best baseline for Ranking: Anchor Points / APP, 0.8428.
1\% & Baseline & 0.6525 (0.6715) & \textbf{0.5937} & \textbf{0.8428} & 0.0344 \\
 & DiagAnchor & \textbf{0.6958} (\textbf{0.6807}) & 0.5732 & 0.6756 & \textbf{0.0311} \\
\addlinespace[3pt]
% 5% best baseline for MAE: tinyBenchmarks IRT + gpIRT, 0.0145.
% 5% best baseline for Full Profile: Anchor Points / APP, 0.6759.
% 5% best baseline for Common Profile: tinyBenchmarks IRT + gpIRT, 0.7172.
% 5% best baseline for Recall@3: Anchor Points / APP, 0.6118.
% 5% best baseline for Ranking: tinyBenchmarks IRT + gpIRT, 0.9099.
5\% & Baseline & 0.6759 (0.7172) & 0.6118 & 0.9099 & \textbf{0.0145} \\
 & DiagAnchor & \textbf{0.8216} (\textbf{0.7759}) & \textbf{0.6297} & \textbf{0.9132} & 0.0203 \\
\addlinespace[3pt]
% 10% best baseline for MAE: tinyBenchmarks IRT + gpIRT, 0.0103.
% 10% best baseline for Full Profile: M-Tag Random, 0.7842.
% 10% best baseline for Common Profile: TailoredBench, 0.7232.
% 10% best baseline for Recall@3: M-Tag Random, 0.6510.
% 10% best baseline for Ranking: tinyBenchmarks IRT + gpIRT, 0.9480.
10\% & Baseline & 0.7842 (0.7232) & 0.6510 & \textbf{0.9480} & \textbf{0.0103} \\
 & DiagAnchor & \textbf{0.8730} (\textbf{0.8326}) & \textbf{0.6991} & 0.9170 & 0.0194 \\
\bottomrule
\end{tabular}
\caption{Fixed-source temporal holdout: DiagAnchor versus the best baseline
for each metric and budget. Bold marks the better mean.}
\label{tab:logo-best-baseline}
\end{table}
\subsection{Ablation Studies}
\label{sec:ablations}

We examine the contributions of anchor selection and calibration
by varying one stage while keeping the other fixed. 
\paragraph{Effect of anchor selection.}
We compare global random sampling (Global Random), M-Tag-stratified random sampling (M-Tag Random),
global anchor selection (Global Anchor), and DiagAnchor selection under
the same local additive calibration. Aggregate-fidelity results
appear in Appendix Figure~\ref{fig:selection-ablation-aggregate}.

\begin{figure*}[t]
    \centering
    \includegraphics[width=\textwidth]{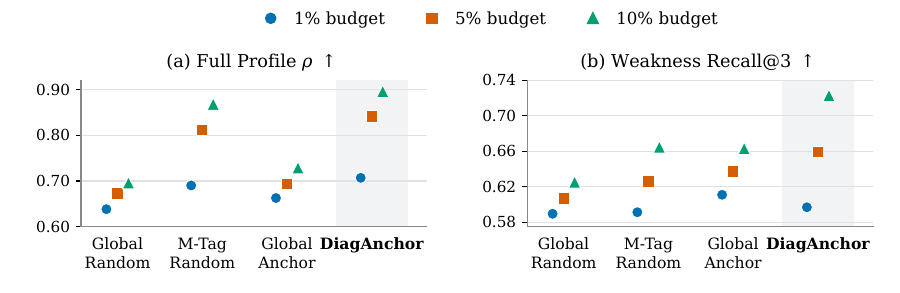}
    \caption{Selection ablation with fixed local additive calibration.
    Shading highlights DiagAnchor selection.}
    \label{fig:selection-ablation}
\end{figure*}

Figure~\ref{fig:selection-ablation} shows that DiagAnchor achieves
the highest mean Full-profile correlation at all three budgets and
the highest mean Recall@3 at 5\% and 10\%.
Relative to Global Random, its Full-profile gain reaches 0.200
at 10\%. This comparison shows that the selection strategy
contributes to diagnostic fidelity even with calibration held fixed.

Comparisons with the other selectors support both parts of the
selection design. Relative to M-Tag Random, DiagAnchor improves
both diagnostic metrics at 5\% and 10\%, supporting the use of
behavioral representatives within capability groups; the Recall@3
gain reaches 5.8 percentage points at 10\%.
Relative to Global Anchor, it also improves both metrics at these
budgets, supporting explicit allocation across the capability
groups used for diagnosis.

The weakness-recovery advantage of stratification grows with the
budget. DiagAnchor trails Global Anchor at 1\%, then exceeds it
by 2.2 and 6.0 percentage points at 5\% and 10\%, respectively.
At 1\%, Global Anchor attains higher Recall@3 than DiagAnchor with
the default exponent (0.611 versus 0.597), but this gap is largely
eliminated with equal group weights ($\beta=0$: 0.610;
Section~\ref{sec:sensitivity}).

%Maybe add MAE and ranking in Appendix

\paragraph{Effect of calibration.}
Table~\ref{tab:calibration-ablation} holds the selected anchors fixed
and replaces local calibration with naive weighted aggregation,
which aggregates queried responses using the weights from anchor
selection. Calibration improves both diagnostic metrics at 5\% and
10\%, and lowers MAE at all three budgets. Its Recall@3 gain increases from 4.8 percentage
points at 10\% to 20.9 at 1\%, where it also roughly halves MAE.
One possible explanation is that source-estimated score differences
provide useful information about unqueried items when direct
observations are sparse. At 1\%, naive aggregation cannot estimate
groups without queried items, so its Full-profile mean covers fewer
cases and is not directly comparable.

\begin{table}[t]
\centering
\footnotesize
\setlength{\tabcolsep}{2pt}
\renewcommand{\arraystretch}{1.12}
\begin{tabular}{@{}clrrrr@{}}
\toprule
Budget & Estimator
& \shortstack{Full Profile\\$\rho\uparrow$}
& R@3 $\uparrow$
& \shortstack{Ranking\\$\rho\uparrow$}
& MAE $\downarrow$ \\
\midrule
1\% & Naive
& $0.5982^{\dagger}$ & 0.3883 & \textbf{0.8397} & 0.0617 \\
& DiagAnchor
& $0.7069$ & \textbf{0.5969} & 0.8356 & \textbf{0.0311} \\
\midrule
5\% & Naive
& 0.7843 & 0.5860 & 0.9050 & 0.0232 \\
& DiagAnchor
& \textbf{0.8414} & \textbf{0.6597} & \textbf{0.9102} & \textbf{0.0198} \\
\midrule
10\% & Naive
& 0.8690 & 0.6747 & 0.9311 & 0.0181 \\
& DiagAnchor
& \textbf{0.8962} & \textbf{0.7231} & \textbf{0.9364} & \textbf{0.0172} \\
\bottomrule
\end{tabular}
\caption{Calibration ablation with DiagAnchor selection held fixed.
Bold marks the better mean, except for Profile at 1\%.}
\label{tab:calibration-ablation}
\vspace{2pt}
\begin{minipage}{\columnwidth}
\footnotesize
$\dagger$ At 1\%, the Profile means use different evaluable cases;
see the accompanying discussion.
\end{minipage}
\end{table}

Stratified selection contributes more to weakness recovery at larger
budgets, whereas calibration helps most when queries are scarce.

\subsection{Sensitivity Analysis}
\label{sec:sensitivity}

\paragraph{Allocation exponent.}
The allocation exponent $\beta$ interpolates between equal group
weights ($\beta=0$) and weights proportional to group size ($\beta=1$).
Across the tested settings, decreasing $\beta$ improves diagnostic
fidelity at every budget; at 5\%, Full-profile correlation rises from
0.771 to 0.883 and Recall@3 from 0.624 to 0.702 as $\beta$ decreases
from 1 to 0, while MAE increases from 0.0186 to 0.0219. At 10\%,
MAE is nearly unchanged across $\beta$. Every tested value yields
higher Full-profile correlation than the strongest baseline at all
three budgets, and the default $\beta=0.5$, fixed a priori, is not
the best diagnostic setting. Full sensitivity results appear in
Appendix~\ref{app:sensitivity-details}.

\paragraph{Source-weighting temperature.}
Diagnostic fidelity is relatively stable as the temperature $\tau$
varies over two orders of magnitude, from 0.01 to 1: Full-profile
correlation changes by at most 0.014 and Recall@3 by at most 2.3
percentage points at any budget. Every tested value exceeds the
strongest baseline on both diagnostic metrics at 5\% and 10\%.
Aggregate fidelity is more sensitive at the low end: the smallest
value, $\tau=0.01$, places the greatest weight on the closest sources
and yields the lowest ranking correlation and highest MAE at each
budget. The effect is most pronounced at 1\% (ranking correlation
0.809 versus 0.836; MAE 0.0355 versus 0.0311, relative to the
default). Further changes are generally smaller for $\tau\geq0.1$.
The default $\tau=0.1$, fixed a priori, lies at the start of this
more stable range.

\paragraph{Source-pool size.}
Halving the source pool, with the target models unchanged, lowers
Full-profile correlation by at most 0.0232 and Recall@3 by at most
2.6 percentage points, while MAE increases by approximately 20--26\%.
With half of the source models, DiagAnchor still exceeds every
full-pool baseline on both diagnostic metrics at 5\% and 10\%.

\section{Conclusion}
\label{sec:conclusion}

Benchmark compression should preserve the capability-level information
needed for model development alongside the scores and rankings used for
model comparison. We formulate diagnostic fidelity through profile
alignment and weakness recovery, using M-Tag as a practical proxy for
capability grouping. The criteria accept evaluator-specified groups and
can therefore be applied with alternative taxonomies. We also propose
DiagAnchor, which combines capability-stratified anchor selection with
target-adaptive local additive calibration. Across eight multilingual
benchmarks, the method shows more consistent diagnostic advantages at
5\% and 10\% query budgets, with trade-offs in aggregate-score and
model-ranking fidelity.
At 1\%, gains remain more dependent on the benchmark and metric.

\section{Limitations}
\label{sec:limitations}

Diagnostic fidelity depends on the grouping scheme and the full benchmark
used as its reference. M-Tag provides an operational proxy for capabilities,
and errors in automatic annotation can affect both the profiles being
evaluated and the allocation of queries. Groups represented by very few
items may also provide an unstable reference for broader capability-level
conclusions. Our metrics therefore measure preservation of the diagnostic
conclusions defined by a particular benchmark and its labels; their
reliability remains dependent on annotation quality and item coverage.

DiagAnchor requires complete responses from historical source models and
assumes that the item relationships captured by these responses transfer
to the target model. The reduced-source and fixed-source temporal-holdout
experiments examine changes in reference-pool size and model availability,
but cover only a limited range of source--target conditions.
Temporal ordering relies on recorded release dates as approximate indicators
of availability and does not directly quantify behavioral distribution shift.
Transfer across substantially different model families or capability
distributions remains to be established.

The diagnostic gains are less consistent under severe query constraints.
At the 1\% budget, relative performance varies across benchmarks and metrics.
Although DiagAnchor has the highest mean profile correlations, its
weakness Recall@3 and model-ranking correlation trail Anchor Points.
The fixed-source temporal-holdout experiment also shows a weakness-recovery
shortfall at this budget. The evidence for improved diagnostic preservation is
more consistent at the 5\% and 10\% budgets; our results do not establish
a uniform advantage at arbitrarily small query budgets.

We report means over target models, splits, and seeds without significance
testing; some differences, particularly at 1\% and on individual
benchmarks, are small. M-Tag validation relies on single-annotator
labels without inter-annotator agreement.

\section*{Ethical Considerations}

This study evaluates model outputs and does not involve human-subject experimentation. Nevertheless, incorrect language or M-Tag assignments can lead to misleading claims about a model's capabilities, particularly for rare or low-resource groups. The single-label taxonomy also compresses potentially overlapping cultural, political, medical, and value-related content into one primary category. Appendix~\ref{app:mtag} reports tagger validation. Compressed subsets should not replace full or human evaluation in high-stakes settings.
% TODO: Confirm whether any benchmark inputs contain personal or sensitive
% data, document proprietary API transmission, and add the venue-required
% disclosure of generative-AI writing assistance if applicable.

\bibliography{custom}
\IfFileExists{acl.sty}{}{%
  % The ACL package sets its bibliography style automatically.
  \bibliographystyle{plainnat}%
}

\clearpage
\appendix

\section{M-Tag Taxonomy, Development, and Validation}
\label{app:mtag}

\paragraph{Taxonomy.}
M-Tag defines 35 level-1 categories and 227 nested level-2 categories.
We use level-1 labels as capability proxies to define diagnostic groups
within each benchmark, pooling items across languages.
% TODO: Provide the official English name and a one-sentence definition for
% each of the 35 first-level tags, plus priority rules for overlapping labels.

\paragraph{Tagger development.} According to the M-Tag taxonomy, we train an automatic tagger in two stages. We first fine-tune Qwen2.5-0.5B-base on approximately 0.8M translated parallel instances spanning 23 languages, yielding a lightweight tagger. We then apply this tagger as a distribution probe over multilingual corpora scraped from Wikipedia to measure the label distribution and counteract the long tail of infrequent tags.

Sampling proceeds per (language, M-Tag) group. Let $N_{\ell,g}$ denote the
number of available instances for language $\ell$ and M-Tag $g$ under the
probe's labelling. We draw $\min(N_{\ell,g}, 300)$ instances per group, taking
all available instances when a group holds fewer than 300. Within each group,
sampling is additionally balanced over instance length to prevent any single
length range from dominating. Sampling above the training quota leaves
headroom for the probe's labelling error: the selected instances are
subsequently relabelled by GPT-5.6, which reassigns a portion of them across
groups, so groups whose size the probe overestimated can still be filled from
real data rather than synthesis.

After relabelling, we retain between 150 and 200 instances per group. Groups
falling below the floor of 150 are supplemented with GPT-5.6-generated data.
The resulting corpus is used to train the final tagger from Qwen3-8B.
\paragraph{Tagger validation.}
We evaluate tagging accuracy on 2,728 in-domain examples and 50
out-of-domain (OOD) examples. Both sets are manually annotated.
Each item is labeled by a single annotator following the M-Tag guidelines.
Inter-annotator agreement was not measured, so these accuracies reflect
agreement with a single reference annotation.
As shown in Table~\ref{tab:tagger}, the level-1 tagger achieves
87.5\% accuracy in-domain and 80.0\% on OOD examples, compared
with 81.5\% and 74.0\%, respectively, for GPT-5.6 prompted with
the same taxonomy. In this work, we use only level-1 tags throughout,
since level-2 tags yield groups too sparse to
support reliable per-group estimation.

\par\medskip
\noindent\begin{minipage}{\columnwidth}
\centering
\footnotesize
\setlength{\tabcolsep}{3pt}
\begin{tabular}{@{}lrrrr@{}}
\toprule
& \multicolumn{2}{c}{Level 1} & \multicolumn{2}{c}{Level 2} \\
\cmidrule(lr){2-3}\cmidrule(lr){4-5}
Model & ID & OOD & ID & OOD \\
\midrule
M-Tag (Qwen3-8B) & 87.5\% & 80.0\% & 79.4\% & 66.0\% \\
GPT-5.6 & 81.5\% & 74.0\% & 71.9\% & 68.0\% \\
\bottomrule
\end{tabular}
\captionof{table}{M-Tag tagging accuracy on human-annotated in-domain (ID)
and out-of-domain (OOD) examples.}
\label{tab:tagger}
\end{minipage}
\par\medskip

\Needspace{6\baselineskip}
\section{DiagAnchor Implementation Details}
\label{app:implementation_details}

\subsection{Budget Allocation}
\label{app:budget-allocation}

For $K\geq H$, each M-Tag group first receives one query. We allocate
the remaining $K-H$ queries according to $w_h=N_h^\beta$, with
additional capacity $N_h-1$ for group $h$. A capacity-constrained
largest-remainder rule fixes saturated groups and redistributes their
excess allocation among groups with remaining capacity. The remaining
quotas are rounded down, and leftover queries are assigned by their
fractional remainders.

For $K<H$, we sequentially sample $K$ distinct groups without
replacement. At each step, the weights are normalized over the
unselected groups; each selected group receives one query.
The resulting integer quotas satisfy
\begin{equation}
    0\leq k_h\leq N_h,\qquad \sum_{h=1}^{H}k_h=K.
\end{equation}
Thus, exactly $\min(K,H)$ groups receive at least one query.

\subsection{Within-Group Anchor Construction}
\label{app:anchor-implementation}

\paragraph{Behavioral clustering.}
Within each group with $k_h>0$, we represent each item
by its score vector across source models. Each source
dimension is standardized over the full item pool using
its sample standard deviation.
Let $U_h$ denote the number of distinct raw source-response
patterns in group $h$. We set the number of clusters to
\begin{equation}
    r_h = \min(k_h,U_h).
\end{equation}

To preserve the contribution of repeated patterns, we
cluster the standardized pattern vectors using weighted
KMeans with the objective
\begin{equation}
    \min_{\{\boldsymbol{\nu}_{hq}\}_{q=1}^{r_h}}
    \sum_{p=1}^{U_h} n_{hp}
    \min_{1 \leq q \leq r_h}
    \left\|
        \mathbf{u}_{hp}-\boldsymbol{\nu}_{hq}
    \right\|_2^2,
\end{equation}
where $\mathbf{u}_{hp}$ is the standardized vector of
pattern $p$, $n_{hp}$ is its item count, and
$\boldsymbol{\nu}_{hq}$ is a cluster center.
When $r_h=U_h$, the distinct response patterns directly
define the clusters. When $r_h=1$, we use the item-count-weighted
mean of the response patterns as the center.

\paragraph{Representative selection.}
We allocate one representative to each cluster and
distribute any remaining quota in proportion to cluster
size, using the same capacity-constrained allocation rule.
Within each cluster, we select the actual item closest
to its center, breaking ties uniformly at random.
If additional representatives are required, we sample
them uniformly without replacement from the remaining
items in that cluster.

Let $S_h$ denote the selected items in group $h$. The final set is
$S_T=\bigcup_h S_h$, with $|S_h|=k_h$ and $|S_T|=K$.
This permits multiple distinct items to represent the same source-response
pattern while satisfying the query budget.

\subsection{Calibration Conventions}
\label{app:calibration-implementation}

For representative matching, source scores are standardized over the
full item pool using the population standard deviation. Selection
uses the sample standard deviation. Constant source dimensions use
a denominator of one; selection also uses one when $N=1$.
Standardized scores define the behavioral geometry, while source
baselines and residual calibration use the original score scale.
Matching ties are resolved by item-key order, and queried items
are mapped to themselves.

Selection assigns each representative a cluster-mass weight:
a cluster containing $M_{hq}$ items and contributing $t_{hq}$
representatives gives each representative weight $M_{hq}/t_{hq}$.
These weights are used by the naive estimator in the calibration
ablation. DiagAnchor instead averages reconstructed item-level
scores equally within each reporting group.

% TODO: Add source-score missingness, exact seed propagation for every method,
% software/hardware details, and repaired top-up behavior.

\subsection{Baseline Implementation}
\label{app:baselines}
\label{app:baseline_details}

All baselines share DiagAnchor's item pools, source/target splits,
selection seeds, and query budgets. For a pool of $N$ items and budget
fraction $b$, the sample size is
$K=\max\{1,\min\{N,\lfloor bN+0.5\rfloor\}\}$. All target responses used
for adaptation count toward the budget, and source fitting and
hyperparameter selection use no unqueried target responses.

\paragraph{Random and M-Tag Random.}
Random samples $K$ items uniformly from the full pool. M-Tag Random uses
DiagAnchor's group allocation ($\beta=0.5$; Section~\ref{sec:diaganchor})
and samples items uniformly within each group. Group estimates are means
of the observed scores, and the overall estimate weights each sampled
item in group $h$ by $N_h/k_h$. Groups without sampled items are left
undefined.

\paragraph{Anchor Points.}
We adapt the Anchor Point Predictor (APP) of \citet{vivek2024anchor} to
non-binary scores by applying a clipped logit transform, which retains
partial credit such as 0.5. Using source models only, $K$ anchors are
selected by FasterPAM with distance $1-|\mathrm{corr}(z_i,z_j)|$,
following the released code. Each item is mapped to the anchor whose
univariate linear regression best predicts it on the source models;
target predictions apply this regression to the anchor's observed score,
and queried items retain their observed scores.

\paragraph{tinyBenchmarks.}
We use the official multidimensional 2PL implementation of
\citet{polo2024tinybenchmarks} with correctness (C) or IRT (I) anchor
embeddings and the pIRT or gpIRT estimator. Anchors are the items nearest
the centroids of KMeans (best of five seeded fits), and the IRT dimension
is selected from $\{2,5,10,15\}$ by validation on held-out source models.
Because the IRT model requires binary outcomes, scores are binarized for
IRT fitting at a threshold chosen to best preserve source-model average
scores; raw scores are retained in the observed-score terms and in
evaluation. For the diagnostic metrics, group estimates average the
item-level IRT probabilities $p_i$ (pIRT) or $\lambda v+(1-\lambda)p_i$
(gpIRT) within each group, where $v$ is the weighted observed mean and
$\lambda$ is the original gpIRT mixing weight; queried items are not
replaced by their observed scores.

\paragraph{TailoredBench.}
Following \citet{yuan2025beyond}, a global coreset of
$G=\min\{10,\max\{1,\lfloor K/2\rfloor\}\}$ items is selected from source
responses by Manhattan-distance CLARA and queried on every target; these
probes count toward the budget. Each target retains its nearest source
models in probe-response space, with the number of native sources set by
a distance threshold computed over the target cohort. A target-specific
$K$-item coreset that includes the probes is then constructed in
native-source space, and each unqueried item is predicted with the
original ratio calibration
$\widetilde{y}_i=(y_{r(i)}+0.5)(a_i+0.5)/(a_{r(i)}+0.5)-0.5$, where
$r(i)$ is the assigned medoid and $a_i$ the native-source mean score.
Overall and group scores average the item-level predictions.

In the main comparison, item-level predictions of DiagAnchor and
TailoredBench are clipped to $[0,1]$; the other methods produce bounded
estimates.

\paragraph{Selection-ablation control.}
Global Anchor selects $K$ anchors by weighted KMeans on standardized
source responses over the full item pool, without stratification. In the
selection ablation, all selectors share DiagAnchor's local additive
calibration.

\section{Additional Experimental Details}
\label{app:experimental-details}
\subsection{Model Pools and Split Protocol}
\label{app:model_splits}

The evaluation pool comprises 45 model configurations across the eight
benchmarks, with 40--42 available per benchmark. Each of the three splits
has disjoint source and target configurations at approximately a 7:3 ratio
and uses five selection seeds, shared across methods.
All models use a consistent inference configuration; the exact settings
and code will be released upon publication.

\subsection{Fixed-Source Temporal Holdout Protocol}
\label{app:temporal-protocol}

The frozen registry contains 39 model configurations with recorded
release dates. We select the ten most recent configurations as targets
and exclude earlier configurations sharing a registry group with any
target. This excludes two earlier configurations, leaving 27 candidate
sources. Availability of benchmark responses determines the final source
pool: 26 for Belebele, INCLUDE, M3Exam, and MGSM; 27 for MMMLU; and 25
for S-AlpacaEval, S-MT-Bench, and SAGE. Within each benchmark, the source
pool is fixed across targets and runs; source weighting or
source-neighborhood selection remains target-adaptive.

Release dates are drawn from official announcements or model repositories,
with provider documentation where necessary. For rolling API aliases
and model-series identifiers, the registry uses first documented public
availability as a proxy; these dates need not identify the exact
snapshots evaluated. The split therefore approximates development-time
evaluation with earlier sources. We reuse the frozen benchmark item
intersections, and selection seeds do not repartition the model pool.

\subsection{Benchmarks and Scoring Protocols}
\label{app:scoring-protocols}

Tables~\ref{tab:benchmarks} and~\ref{tab:benchmark-properties} summarize
the evaluated item pools. All recorded outcomes are mapped to scores in
$\{0,0.5,1\}$. Evaluated-subset percentages describe these pools and
are separate from the compression budgets.
Item-level scoring follows the GaoYao evaluation protocol
\citep{liu2026gaoyao}. S-AlpacaEval and S-MT-Bench are GaoYao's
multilingual extensions of AlpacaEval \citep{li2023alpacaeval} and
MT-Bench \citep{zheng2023judging}, respectively.

For S-AlpacaEval and S-MT-Bench, each response is compared with a
reference response generated by Qwen3-235B-A22B-Instruct-2507, with
DeepSeek-V3.1 serving as the judge in both presentation orders.
% Reference version follows the Qwen3-235B-A22B mapping in GaoYao Table 7.
% Confirm against the actual generation registry if it differs from that release.
Because the reference and judge models also belong to the evaluated
pool, their full-benchmark scores may be biased. All compression
methods share these benchmark scores as their evaluation reference;
this does not eliminate potential reference or judge bias in comparisons.

For S-AlpacaEval, S-MT-Bench, and SAGE, we did not find explicit
dataset-specific license statements in the source publications.
We do not redistribute any datasets as part of this work.

\begin{table*}[!tp]
\centering
\footnotesize
\setlength{\tabcolsep}{3pt}
\renewcommand{\arraystretch}{1.15}
\begin{tabular}{@{}lrrrrrl@{}}
\toprule
Benchmark & Items & Languages & L1 M-Tags & Models & Source/target & Data license \\
\midrule
Belebele \citep{bandarkar2024belebele} & 18,900 & 21 & 34 & 41 & 29/12 & CC BY-SA 4.0 \\
INCLUDE \citep{romanou2025include} & 9,117 & 18 & 35 & 41 & 29/12 & Apache 2.0 \\
M3Exam \citep{zhang2023m3exam} & 39,105 & 21 & 33 & 41 & 29/12 & CC BY-NC-SA \\
MGSM \citep{shi2023multilingual} & 2,750 & 11 & 27 & 41 & 29/12 & CC BY-SA 4.0 \\
MMMLU \citep{openai2024mmmlu} & 19,639 & 14 & 35 & 42 & 29/13 & MIT \\
S-AlpacaEval \citep{liu2026gaoyao} & 4,580 & 19 & 35 & 42 & 29/13 & Not stated \\
S-MT-Bench \citep{liu2026gaoyao} & 895 & 19 & 28 & 40 & 28/12 & Not stated \\
SAGE \citep{guo2025sage} & 571 & 2 & 16 & 42 & 29/13 & Not stated \\
\bottomrule
\end{tabular}
\caption{Benchmark statistics.
L1 M-Tags: distinct first-level labels; Models: unique model count;
Source/target: model counts per split.}
\label{tab:benchmarks}
% License sources checked 2026-10-13:
% Belebele: https://github.com/facebookresearch/belebele (evaluation data).
% INCLUDE: https://huggingface.co/datasets/CohereLabs/include-base-44
% M3Exam: original NeurIPS 2023 paper, Appendix A.1.3 (no version specified).
% MMMLU: https://huggingface.co/datasets/openai/MMMLU
% MGSM: https://huggingface.co/datasets/juletxara/mgsm (HF dataset license metadata).
% No dataset-specific license statement was found in the cited GaoYao/SAGE
% publications. This describes the available documentation and does not
% assert author permission or assign a license to these datasets.
\end{table*}

\begin{table*}[!tp]
\centering
\small
\renewcommand{\arraystretch}{1.15}
\setlength{\tabcolsep}{3pt}
\begin{tabular}{lll}
\toprule
Benchmark & Task format & Evaluated subset \\
\midrule
Belebele       & MCQ                 & Full; 21 languages \\
INCLUDE        & MCQ                 & Full; 18 languages \\
M3Exam         & MCQ, partly visual  & Full; 21 languages \\
MGSM           & Integer math QA     & Full; 11 languages \\
MMMLU          & MCQ                 & 10\%; 14 languages \\
S-AlpacaEval   & Open-ended QA       & 30\%; 19 languages \\
S-MT-Bench     & Multi-turn QA       & 30\%; 19 languages \\
SAGE          & MCQ/T--F/open QA    & Full; 2 languages \\
\bottomrule
\end{tabular}
\caption{Benchmark task formats and evaluated subsets.
Scoring follows the GaoYao protocol.}
\label{tab:benchmark-properties}
\end{table*}

\subsection{Additional tinyBenchmarks Results}
\label{app:pirt-results}
Table~\ref{tab:pirt-results} reports the pIRT variants omitted from the
main table. The evaluation protocol is unchanged.

\begin{table*}[!tp]
\centering
\small
\setlength{\tabcolsep}{7pt}
\renewcommand{\arraystretch}{1.08}
\begin{tabular}{@{}clrrrr@{}}
\toprule
Budget & Emb. & Profile $\rho\uparrow$ & R@3$\uparrow$ & Ranking$\uparrow$ & MAE$\downarrow$ \\
& & Full (Common) & & & \\
\midrule
1\% & C & 0.6384 (0.6358) & 0.5805 & 0.8565 & 0.0557 \\
 & I & 0.6573 (0.6174) & 0.5814 & 0.8375 & 0.0497 \\
5\% & C & 0.6848 (0.7499) & 0.6211 & 0.9150 & 0.0441 \\
 & I & 0.6852 (0.7440) & 0.6160 & 0.8978 & 0.0426 \\
10\% & C & 0.7084 (0.7745) & 0.6251 & 0.9495 & 0.0363 \\
 & I & 0.6981 (0.7658) & 0.6233 & 0.9298 & 0.0355 \\
\bottomrule
\end{tabular}
\caption{tinyBenchmarks with pIRT estimation.
C/I: correctness/IRT embeddings.}
\label{tab:pirt-results}
\end{table*}

\section{Per-benchmark Diagnostic Results}
\label{app:diagnostic-results}

Tables~\ref{tab:per-benchmark-mtag-1} and~\ref{tab:per-benchmark-mtag-2}
report per-benchmark diagnostic fidelity for all methods, including the
pIRT variants. Profile cells use the same Full (Common) format; R@3 denotes
weakness Recall@3 and NA denotes an undefined mean.

DiagAnchor leads under Full Profile on all eight benchmarks at 5\% and
10\%. Under Common Profile, it leads on five benchmarks at 5\% and seven
at 10\%; tinyBenchmarks leads on INCLUDE, MGSM, and MMMLU at 5\%, and
TailoredBench leads on SAGE at 10\%. For Recall@3, DiagAnchor leads on
four benchmarks at 5\% and all eight at 10\%. At 1\%, it leads on four
benchmarks under Full Profile, three under Common Profile, and one for
Recall@3.

\begin{table*}[!tp]
\centering
\footnotesize
\setlength{\tabcolsep}{2.5pt}
\renewcommand{\arraystretch}{1.08}
\begin{tabular}{@{}llrrrrrr@{}}
\toprule
Benchmark & Method & \multicolumn{2}{c}{1\%}
& \multicolumn{2}{c}{5\%} & \multicolumn{2}{c}{10\%} \\
\cmidrule(lr){3-4}\cmidrule(lr){5-6}\cmidrule(lr){7-8}
& & \shortstack{Profile $\rho$\\Full (Common)} & R@3
& \shortstack{Profile $\rho$\\Full (Common)} & R@3
& \shortstack{Profile $\rho$\\Full (Common)} & R@3 \\
\midrule
\multirow{9}{*}{Belebele} & Random & \textsc{na} (0.2896) & 0.2204 & \textsc{na} (0.4345) & 0.3537 & \textsc{na} (0.5775) & 0.4185 \\
 & M-Tag Random & 0.3762 (0.2904) & 0.3204 & 0.6771 (0.5824) & 0.5093 & 0.8592 (0.7739) & 0.6241 \\
 & Anchor Points (APP) & 0.8078 (0.8503) & 0.6407 & 0.8153 (0.8562) & 0.6481 & 0.8141 (0.8526) & 0.6556 \\
 & TailoredBench & 0.8084 (0.8479) & 0.6167 & 0.7945 (0.8323) & 0.6519 & 0.8283 (0.8663) & 0.6611 \\
 & tinyB. (C, pIRT) & 0.8373 (\textbf{0.8747}) & \textbf{0.6778} & 0.8499 (0.8857) & 0.7148 & 0.8498 (0.8852) & 0.7019 \\
 & tinyB. (C, gpIRT) & 0.8373 (\textbf{0.8747}) & \textbf{0.6778} & 0.8499 (0.8857) & 0.7148 & 0.8498 (0.8852) & 0.7019 \\
 & tinyB. (I, pIRT) & 0.8297 (0.8707) & \textbf{0.6778} & 0.8469 (0.8873) & 0.6852 & 0.8513 (0.8871) & 0.6889 \\
 & tinyB. (I, gpIRT) & 0.8297 (0.8707) & \textbf{0.6778} & 0.8469 (0.8873) & 0.6852 & 0.8513 (0.8871) & 0.6889 \\
 & \textbf{DiagAnchor} & \textbf{0.8790} (0.8629) & 0.6519 & \textbf{0.9288} (\textbf{0.9038}) & \textbf{0.7685} & \textbf{0.9542} (\textbf{0.9305}) & \textbf{0.8222} \\
\midrule
\multirow{9}{*}{INCLUDE} & Random & \textsc{na} (0.2531) & 0.1556 & \textsc{na} (0.4075) & 0.3778 & \textsc{na} (0.5249) & 0.3611 \\
 & M-Tag Random & 0.3122 (0.2032) & 0.2148 & 0.6143 (0.5451) & 0.3611 & 0.7786 (0.7129) & 0.5074 \\
 & Anchor Points (APP) & 0.7805 (0.8772) & 0.5759 & 0.7899 (0.8834) & 0.5833 & 0.7925 (0.8787) & 0.6019 \\
 & TailoredBench & 0.8008 (0.9028) & 0.5704 & 0.7843 (0.8608) & 0.5833 & 0.7865 (0.8560) & 0.5870 \\
 & tinyB. (C, pIRT) & 0.8298 (\textbf{0.9149}) & \textbf{0.6370} & 0.8427 (0.9140) & \textbf{0.6611} & 0.8430 (0.9123) & 0.6370 \\
 & tinyB. (C, gpIRT) & 0.8298 (\textbf{0.9149}) & \textbf{0.6370} & 0.8427 (0.9140) & \textbf{0.6611} & 0.8430 (0.9123) & 0.6370 \\
 & tinyB. (I, pIRT) & \textbf{0.8331} (0.9084) & 0.6296 & 0.8451 (\textbf{0.9152}) & 0.6463 & 0.8441 (0.9112) & 0.6389 \\
 & tinyB. (I, gpIRT) & \textbf{0.8331} (0.9084) & 0.6296 & 0.8451 (\textbf{0.9152}) & 0.6463 & 0.8441 (0.9112) & 0.6389 \\
 & \textbf{DiagAnchor} & 0.8249 (0.8825) & 0.5556 & \textbf{0.8936} (0.8816) & 0.5907 & \textbf{0.9378} (\textbf{0.9177}) & \textbf{0.6556} \\
\midrule
\multirow{9}{*}{M3Exam} & Random & \textsc{na} (0.3194) & 0.2296 & \textsc{na} (0.4938) & 0.3889 & \textsc{na} (0.5853) & 0.4370 \\
 & M-Tag Random & 0.5735 (0.5457) & 0.4593 & 0.8958 (0.8417) & 0.6667 & 0.9639 (0.9365) & 0.7907 \\
 & Anchor Points (APP) & 0.7554 (0.8385) & 0.6037 & 0.7439 (0.8135) & 0.6204 & 0.7536 (0.8163) & 0.6204 \\
 & TailoredBench & 0.7422 (0.8447) & 0.5741 & 0.7476 (0.8283) & 0.6130 & 0.7595 (0.8339) & 0.6000 \\
 & tinyB. (C, pIRT) & 0.7859 (0.8499) & 0.5704 & 0.7875 (0.8355) & 0.5741 & 0.7841 (0.8331) & 0.5741 \\
 & tinyB. (C, gpIRT) & 0.7859 (0.8499) & 0.5704 & 0.7875 (0.8355) & 0.5741 & 0.7841 (0.8331) & 0.5741 \\
 & tinyB. (I, pIRT) & 0.7835 (0.8437) & 0.5685 & 0.7923 (0.8397) & 0.5741 & 0.7928 (0.8412) & 0.5741 \\
 & tinyB. (I, gpIRT) & 0.7835 (0.8437) & 0.5685 & 0.7923 (0.8397) & 0.5741 & 0.7928 (0.8412) & 0.5741 \\
 & \textbf{DiagAnchor} & \textbf{0.8497} (\textbf{0.8706}) & \textbf{0.6407} & \textbf{0.9550} (\textbf{0.9282}) & \textbf{0.7778} & \textbf{0.9849} (\textbf{0.9743}) & \textbf{0.9056} \\
\midrule
\multirow{9}{*}{MGSM} & Random & \textsc{na} (0.1622) & 0.1616 & \textsc{na} (0.3505) & 0.3755 & \textsc{na} (0.3835) & 0.4542 \\
 & M-Tag Random & 0.3644 (0.2226) & 0.3347 & 0.6263 (0.4161) & 0.5019 & 0.8203 (0.6224) & 0.5935 \\
 & Anchor Points (APP) & 0.6384 (0.6094) & 0.6750 & 0.6499 (0.7075) & 0.6759 & 0.6496 (0.7428) & 0.7125 \\
 & TailoredBench & 0.6316 (0.6283) & 0.6565 & 0.6624 (0.7188) & 0.6713 & 0.7017 (0.7627) & 0.7051 \\
 & tinyB. (C, pIRT) & 0.6646 (\textbf{0.7635}) & 0.7093 & 0.6842 (\textbf{0.8001}) & \textbf{0.7204} & 0.6797 (0.7924) & 0.6981 \\
 & tinyB. (C, gpIRT) & 0.6646 (\textbf{0.7635}) & 0.7093 & 0.6842 (\textbf{0.8001}) & \textbf{0.7204} & 0.6797 (0.7924) & 0.6981 \\
 & tinyB. (I, pIRT) & 0.6676 (0.7316) & \textbf{0.7259} & 0.6766 (0.7959) & 0.7130 & 0.6740 (0.7914) & 0.7148 \\
 & tinyB. (I, gpIRT) & 0.6676 (0.7316) & \textbf{0.7259} & 0.6766 (0.7959) & 0.7130 & 0.6740 (0.7914) & 0.7148 \\
 & \textbf{DiagAnchor} & \textbf{0.6869} (0.6735) & 0.6648 & \textbf{0.8717} (0.7607) & 0.6810 & \textbf{0.9345} (\textbf{0.8421}) & \textbf{0.7380} \\
\bottomrule
\end{tabular}
\caption{Per-benchmark diagnostic fidelity (part 1 of 2).
Bold marks the best mean for each budget, metric, and profile protocol.
tinyB.: tinyBenchmarks; C/I: correctness/IRT embeddings; APP: Anchor Points.}
\label{tab:per-benchmark-mtag-1}
\end{table*}

\begin{table*}[!tp]
\centering
\footnotesize
\setlength{\tabcolsep}{2.5pt}
\renewcommand{\arraystretch}{1.08}
\begin{tabular}{@{}llrrrrrr@{}}
\toprule
Benchmark & Method & \multicolumn{2}{c}{1\%}
& \multicolumn{2}{c}{5\%} & \multicolumn{2}{c}{10\%} \\
\cmidrule(lr){3-4}\cmidrule(lr){5-6}\cmidrule(lr){7-8}
& & \shortstack{Profile $\rho$\\Full (Common)} & R@3
& \shortstack{Profile $\rho$\\Full (Common)} & R@3
& \shortstack{Profile $\rho$\\Full (Common)} & R@3 \\
\midrule
\multirow{9}{*}{MMMLU} & Random & \textsc{na} (0.3691) & 0.2957 & \textsc{na} (0.5424) & 0.2974 & \textsc{na} (0.6043) & 0.3846 \\
 & M-Tag Random & 0.4242 (0.4369) & 0.2547 & 0.7131 (0.6922) & 0.4735 & 0.8595 (0.8284) & 0.6735 \\
 & Anchor Points (APP) & 0.7492 (0.8741) & \textbf{0.7504} & 0.7578 (0.8581) & \textbf{0.7744} & 0.7649 (0.8488) & 0.7590 \\
 & TailoredBench & 0.7455 (0.8681) & 0.7145 & 0.7420 (0.8390) & 0.7538 & 0.7614 (0.8451) & 0.7453 \\
 & tinyB. (C, pIRT) & \textbf{0.8153} (0.9047) & 0.7333 & 0.8252 (0.8963) & 0.7385 & 0.8248 (0.8910) & 0.7487 \\
 & tinyB. (C, gpIRT) & \textbf{0.8153} (0.9047) & 0.7333 & 0.8252 (0.8963) & 0.7385 & 0.8248 (0.8910) & 0.7487 \\
 & tinyB. (I, pIRT) & 0.8078 (\textbf{0.9086}) & 0.7316 & 0.8259 (\textbf{0.8977}) & 0.7368 & 0.8271 (0.8931) & 0.7385 \\
 & tinyB. (I, gpIRT) & 0.8078 (\textbf{0.9086}) & 0.7316 & 0.8259 (\textbf{0.8977}) & 0.7368 & 0.8271 (0.8931) & 0.7385 \\
 & \textbf{DiagAnchor} & 0.7826 (0.8726) & 0.7299 & \textbf{0.8799} (0.8918) & 0.7436 & \textbf{0.9368} (\textbf{0.9293}) & \textbf{0.7607} \\
\midrule
\multirow{9}{*}{S-AlpacaEval} & Random & \textsc{na} (0.2819) & 0.1507 & \textsc{na} (0.3926) & 0.1941 & \textsc{na} (0.4897) & 0.2532 \\
 & M-Tag Random & 0.2631 (0.2327) & 0.1516 & 0.4834 (0.4346) & 0.2404 & 0.6362 (0.5891) & 0.3318 \\
 & Anchor Points (APP) & 0.4137 (0.4594) & 0.2547 & 0.4298 (0.4987) & 0.2946 & 0.4963 (0.5425) & 0.3414 \\
 & TailoredBench & 0.4202 (0.5055) & 0.2923 & 0.4459 (0.5041) & 0.2889 & 0.5003 (0.5711) & 0.3414 \\
 & tinyB. (C, pIRT) & 0.5087 (0.5445) & 0.3163 & 0.5421 (0.5907) & 0.3041 & 0.5431 (0.5977) & 0.3092 \\
 & tinyB. (C, gpIRT) & 0.5085 (0.5444) & \textbf{0.3180} & 0.5423 (0.5907) & 0.3041 & 0.5429 (0.5977) & 0.3092 \\
 & tinyB. (I, pIRT) & \textbf{0.5183} (0.5546) & 0.3114 & 0.5324 (0.5828) & 0.3106 & 0.5350 (0.5920) & 0.2958 \\
 & tinyB. (I, gpIRT) & 0.5182 (\textbf{0.5547}) & 0.3102 & 0.5324 (0.5828) & 0.3106 & 0.5357 (0.5920) & 0.2958 \\
 & \textbf{DiagAnchor} & 0.4551 (0.5053) & 0.2833 & \textbf{0.6222} (\textbf{0.6097}) & \textbf{0.3551} & \textbf{0.6990} (\textbf{0.6792}) & \textbf{0.4655} \\
\midrule
\multirow{9}{*}{S-MT-Bench} & Random & \textsc{na} (0.3077) & 0.2198 & \textsc{na} (0.4307) & 0.4007 & \textsc{na} (0.4460) & 0.4105 \\
 & M-Tag Random & \textsc{na} (0.1761) & 0.2224 & 0.5501 (0.3806) & 0.3809 & 0.6785 (0.4692) & 0.4296 \\
 & Anchor Points (APP) & 0.4840 (0.3800) & \textbf{0.6138} & 0.5258 (0.6092) & 0.6107 & 0.5243 (0.6464) & 0.6222 \\
 & TailoredBench & 0.4752 (0.3189) & 0.5669 & 0.4983 (0.5933) & 0.6355 & 0.5379 (0.6464) & 0.6616 \\
 & tinyB. (C, pIRT) & 0.1936 (0.1687) & 0.3593 & 0.4198 (0.5021) & 0.5622 & 0.4917 (0.5887) & 0.5910 \\
 & tinyB. (C, gpIRT) & 0.1934 (0.1687) & 0.3593 & 0.4197 (0.5021) & 0.5622 & 0.4917 (0.5887) & 0.5910 \\
 & tinyB. (I, pIRT) & 0.2421 (0.1520) & 0.4000 & 0.4088 (0.4832) & 0.5567 & 0.4574 (0.5486) & 0.5885 \\
 & tinyB. (I, gpIRT) & 0.2420 (0.1520) & 0.4000 & 0.4089 (0.4832) & 0.5567 & 0.4574 (0.5486) & 0.5885 \\
 & \textbf{DiagAnchor} & \textbf{0.5255} (\textbf{0.4433}) & 0.5864 & \textbf{0.6763} (\textbf{0.6229}) & \textbf{0.6469} & \textbf{0.7893} (\textbf{0.6562}) & \textbf{0.6704} \\
\midrule
\multirow{9}{*}{SAGE} & Random & \textsc{na} (0.2619) & 0.4333 & \textsc{na} (0.1712) & 0.5295 & \textsc{na} (0.3342) & 0.5932 \\
 & M-Tag Random & \textsc{na} (0.2498) & 0.4547 & 0.4859 (0.1750) & 0.5235 & 0.6530 (0.3859) & 0.6038 \\
 & Anchor Points (APP) & \textbf{0.7793} (0.6014) & \textbf{0.7410} & 0.7476 (0.5864) & \textbf{0.7581} & 0.7448 (0.6127) & 0.7547 \\
 & TailoredBench & 0.6521 (0.6458) & 0.7325 & 0.7084 (0.6129) & 0.7256 & 0.7720 (\textbf{0.7039}) & 0.7496 \\
 & tinyB. (C, pIRT) & 0.4722 (0.0653) & 0.6402 & 0.5269 (0.5746) & 0.6936 & 0.6510 (0.6959) & 0.7410 \\
 & tinyB. (C, gpIRT) & 0.4722 (0.0653) & 0.6402 & 0.5269 (0.5746) & 0.6936 & 0.6510 (0.6959) & 0.7410 \\
 & tinyB. (I, pIRT) & 0.5761 (-0.0306) & 0.6060 & 0.5533 (0.5502) & 0.7051 & 0.6035 (0.6619) & 0.7466 \\
 & tinyB. (I, gpIRT) & 0.5761 (-0.0306) & 0.6060 & 0.5532 (0.5502) & 0.7051 & 0.6035 (0.6619) & 0.7466 \\
 & \textbf{DiagAnchor} & 0.6514 (\textbf{0.6514}) & 0.6624 & \textbf{0.9035} (\textbf{0.6189}) & 0.7137 & \textbf{0.9330} (0.6858) & \textbf{0.7667} \\
\bottomrule
\end{tabular}
\caption{Per-benchmark diagnostic fidelity (part 2 of 2).
Notation follows Table~\ref{tab:per-benchmark-mtag-1}.}
\label{tab:per-benchmark-mtag-2}
\end{table*}

\section{Additional Ablation and Sensitivity Results}
\label{app:additional-analyses}
\subsection{Aggregate Fidelity in the Selection Ablation}
\label{app:selection-aggregate}

\begin{figure*}[!tp]
\centering
    \includegraphics[width=\textwidth]{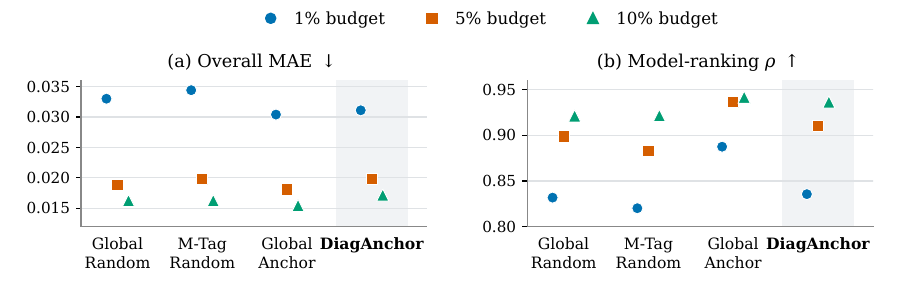}
    \caption{Aggregate fidelity in the selection ablation.
    Shading highlights DiagAnchor selection.}
    \label{fig:selection-ablation-aggregate}
\end{figure*}

Figure~\ref{fig:selection-ablation-aggregate} reports overall score
MAE and model-ranking correlation for the selection ablation,
using the same selectors and calibration as
Figure~\ref{fig:selection-ablation}.

\subsection{Sensitivity to Source Models and Hyperparameters}
\label{app:sensitivity-details}

All three sensitivity analyses report Full-profile correlation
and weakness Recall@3.
% TODO: Verify the evaluation support and aggregation for the provisionally
% standardized 1% Ranking mean (0.8356) against the underlying evaluation rows.

\subsubsection{Source-pool Size}
We assess sensitivity to the availability of reference-model
responses by reducing the source pool to half its original size
while keeping the target-model query budget unchanged.
Table~\ref{tab:source-pool-sensitivity} compares this reduced-pool
setting with the default configuration.

% Source: author-verified corrected Full-profile report, 2026-10-11.
% TODO: Specify how the source subset is selected and whether
% both selection and calibration are rerun with the reduced pool.

\par\medskip
\noindent\begin{minipage}{\columnwidth}
\centering
\footnotesize
\setlength{\tabcolsep}{3pt}
\renewcommand{\arraystretch}{1.12}
\begin{tabular}{@{}clrrrr@{}}
\toprule
Budget & Sources
& \shortstack{Full Profile\\$\rho\uparrow$}
& \shortstack{R@3\\$\uparrow$}
& \shortstack{Ranking\\$\rho\uparrow$}
& \shortstack{MAE\\$\downarrow$} \\
\midrule
1\% & Full
& 0.7069 & 0.5969 & 0.8356 & 0.0311 \\
& Half
& 0.6837 & 0.5827 & 0.8335 & 0.0390 \\
\midrule
5\% & Full
& 0.8414 & 0.6597 & 0.9102 & 0.0198 \\
& Half
& 0.8249 & 0.6427 & 0.9020 & 0.0250 \\
\midrule
10\% & Full
& 0.8962 & 0.7231 & 0.9364 & 0.0172 \\
& Half
& 0.8817 & 0.6971 & 0.9173 & 0.0207 \\
\bottomrule
\end{tabular}
\captionof{table}{Sensitivity to source-pool size.
Full denotes the default pool; Half retains half of its models.}
\label{tab:source-pool-sensitivity}
\end{minipage}
\par\medskip

Reducing the source pool lowers average diagnostic fidelity at all
three query budgets. Full-profile correlation decreases by 0.0232,
0.0165, and 0.0145 at 1\%, 5\%, and 10\%, respectively,
while Recall@3 decreases by 1.42, 1.70, and 2.60 percentage
points. At 10\%, the reduced pool yields a Full-profile correlation
of 0.8817 and a Recall@3 of 0.6971, compared with 0.8962 and
0.7231 under the full pool. M-Tag query coverage remains unchanged
at each budget, including complete coverage at 5\% and 10\%.
Thus, coverage alone does not account for diagnostic quality;
the available source responses also affect profile reconstruction.

Aggregate fidelity also declines with fewer sources. Model-ranking
correlation decreases by 0.0021, 0.0082, and 0.0191 at the three
budgets, while MAE increases from 0.0311 to 0.0390 at 1\%,
from 0.0198 to 0.0250 at 5\%, and from 0.0172 to 0.0207
at 10\%. DiagAnchor therefore benefits from a larger reference
pool even when capability coverage is preserved. The Full-profile
gap narrows as the query budget increases, although the same
pattern does not hold for weakness recovery or model rankings.

% Hyperparameter sensitivity dashboard.

% Requires graphicx; copy figures/tau_beta_sensitivity.pdf into the paper project.
% Both beta and tau use the final [0,1]-clipped runs.

\begin{figure*}[!tp]
\centering
    \includegraphics[width=\textwidth]{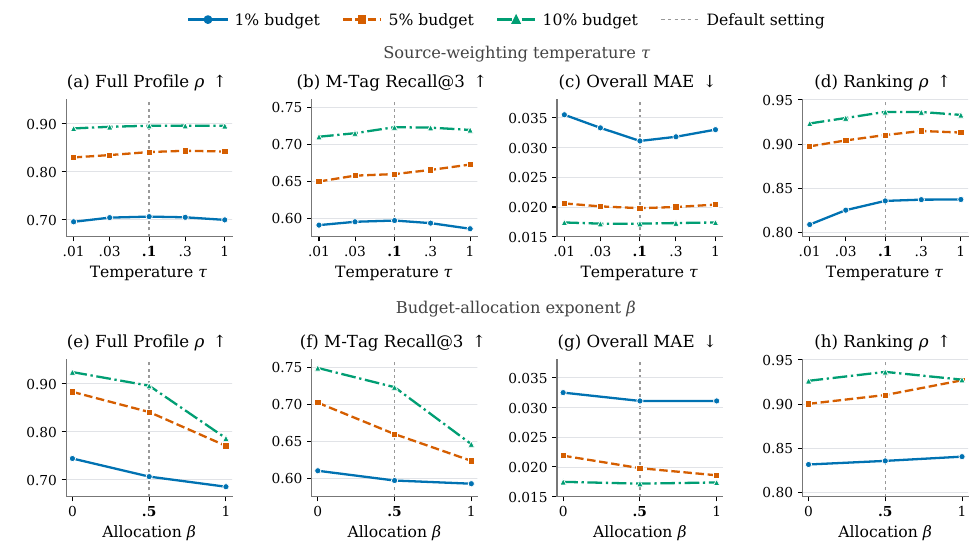}
    \caption{Sensitivity to temperature $\tau$ (top) and allocation exponent
    $\beta$ (bottom). Gray lines mark the defaults.}
    \label{fig:hyperparameter-sensitivity}
\end{figure*}

\subsubsection{Source-weighting Temperature $\tau$}
Figure~\ref{fig:hyperparameter-sensitivity}(a--d) varies $\tau$
over $\{0.01,0.03,0.1,0.3,1\}$ with $\beta=0.5$.
Smaller temperatures concentrate weight on source
models that agree most closely with the target's queried responses;
larger temperatures produce flatter weights.
At the 5\% budget, increasing $\tau$ from 0.01 to 0.3 improves
Full-profile correlation from 0.8302 to 0.8440, followed by a slight
decrease to 0.8425 at $\tau=1$. Recall@3 increases monotonically
over the tested range, from 0.6497 to 0.6726.
The aggregate metrics favor intermediate temperatures:
MAE is lowest at $\tau=0.1$ (0.0198), while model-ranking
correlation is highest at $\tau=0.3$ (0.9148).

The diagnostic benefit of increasing temperature varies with the budget.
At 1\%, both metrics peak at the default $\tau=0.1$, with Full Profile
of 0.7069 and Recall@3 of 0.5969. At 10\%, Full Profile reaches
0.8964 at $\tau=0.3$, only 0.0002 above the default, while Recall@3
is highest at $\tau=0.1$ (0.7231). A broader source mixture therefore
benefits weakness recovery at 5\%, but provides no consistent advantage
across budgets. We retain $\tau=0.1$ for the main comparison.

\subsubsection{Budget-allocation Exponent $\beta$}
Figure~\ref{fig:hyperparameter-sensitivity}(e--h) varies $\beta$
over $\{0,0.5,1\}$ with $\tau=0.1$.
Under the same coverage and capacity constraints, $\beta=0$ assigns
equal allocation weights to capability groups, whereas $\beta=1$
assigns weights proportional to their sizes. Both Full-profile
correlation and Recall@3 decrease across these three settings at
every query budget. From $\beta=0$ to $\beta=1$, Full Profile
falls from 0.7446 to 0.6858 at 1\%, from 0.8834 to 0.7709
at 5\%, and from 0.9246 to 0.7868 at 10\%. The corresponding
Recall@3 values decrease from 0.6101 to 0.5926, from 0.7021
to 0.6235, and from 0.7491 to 0.6461.

These results favor more balanced allocation for capability diagnosis.
All three settings achieve complete M-Tag query coverage at 5\%
and 10\%, so the diagnostic differences at these budgets arise
despite identical group coverage. This is consistent with allocating
more observations to small groups improving their representation
in diagnostics that treat groups as the units of comparison.
The results support an allocation trade-off, without establishing
that equal weights are optimal beyond the tested settings.

Aggregate fidelity shows a different preference. At 5\%, increasing
$\beta$ from 0 to 1 reduces MAE from 0.0219 to 0.0186 and
increases model-ranking correlation from 0.9002 to 0.9269.
This pattern does not extend to every budget: at 10\%, the default
$\beta=0.5$ gives both the lowest MAE (0.0172) and the highest
ranking correlation (0.9364) among the tested settings. We retain
the default $\beta=0.5$ in the main comparison and report the
diagnostic gains from equal weighting as a sensitivity result.

\twocolumn[{%
\begin{minipage}{\textwidth}
\centering
\small
\setlength{\tabcolsep}{6pt}
\renewcommand{\arraystretch}{1.12}
\begin{tabular}{@{}lrrrrrr@{}}
\toprule
& \multicolumn{2}{c}{1\%} & \multicolumn{2}{c}{5\%} & \multicolumn{2}{c}{10\%} \\
\cmidrule(lr){2-3}\cmidrule(lr){4-5}\cmidrule(lr){6-7}
Method & Profile $\rho\uparrow$ & R@3 $\uparrow$
& Profile $\rho\uparrow$ & R@3 $\uparrow$
& Profile $\rho\uparrow$ & R@3 $\uparrow$ \\
\midrule
Random & 0.3025 & 0.3101 & 0.4883 & 0.4719 & 0.5838 & 0.5224 \\
M-Tag Random & 0.2806 & 0.3062 & 0.4229 & 0.4188 & 0.5344 & 0.4870 \\
Anchor Points & 0.6601 & 0.5817 & 0.7019 & 0.5991 & 0.7168 & 0.6286 \\
TailoredBench & \textbf{0.6850} & \textbf{0.6048} & 0.7141 & 0.6056 & 0.7434 & 0.6299 \\
tinyBenchmarks (C, pIRT) & 0.6375 & 0.5829 & 0.7132 & 0.6105 & 0.7226 & 0.6144 \\
tinyBenchmarks (C, gpIRT) & 0.6375 & 0.5829 & 0.7132 & 0.6105 & 0.7226 & 0.6144 \\
tinyBenchmarks (I, pIRT) & 0.6835 & 0.5873 & 0.7129 & 0.6112 & 0.7215 & 0.6131 \\
tinyBenchmarks (I, gpIRT) & 0.6835 & 0.5873 & 0.7129 & 0.6112 & 0.7215 & 0.6131 \\
\midrule
\textbf{DiagAnchor (ours)} & 0.6611 & 0.5861
& \textbf{0.7232} & \textbf{0.6262} & \textbf{0.7470} & \textbf{0.6410} \\
\bottomrule
\end{tabular}
\captionof{table}{Language-axis diagnostic fidelity. Profile uses common language
groups; bold marks the best value in each column.}
\label{tab:language-diagnostics}
\end{minipage}
\vspace{8pt}
}]

\section{Diagnostic Fidelity along the Language Axis}
\label{app:language-diagnostics}

We additionally group items by language and evaluate profile alignment
and weakness recovery using the same selected items and predictions.
Neither selection nor calibration is modified for this analysis.
Table~\ref{tab:language-diagnostics} reports Common Profile correlation
and Recall@3, averaged equally over the seven contributing benchmarks.
\newpage
Language Recall@3 excludes SAGE because it contains only two languages.

DiagAnchor achieves the highest mean values on both language-level
metrics at 5\% and 10\%, although it allocates queries only across
capability groups. At 1\%, it trails TailoredBench on both metrics
and the IRT-embedding variants of tinyBenchmarks on profile alignment.
Thus, the evidence for preserving diagnosis beyond the selection axis
is stronger at the two larger budgets.

\section{Use of Generative AI}
We used generative AI tools to assist with code development and manuscript polishing, including improvements to clarity and wording. The authors take responsibility for the final code, analyses, and manuscript.
\end{document}